%% file: paper.tex
\documentclass[11pt]{article}
\usepackage{stylesheet/coling2018}
\usepackage{times}
\usepackage{url}
\usepackage{latexsym}

\usepackage{mathtools}
\usepackage{multirow}
\usepackage{amsmath}
\usepackage{amssymb}
\usepackage{booktabs}
\usepackage{subcaption}
\usepackage{wrapfig}
\usepackage{color}
\usepackage[table]{xcolor}
\definecolor{lightred}{RGB}{255,110,71}
\definecolor{lightblue}{RGB}{110,130,255}

\newcommand{\thistask}{\textsc{ERM}}

\usepackage{pifont}
\newcommand{\cmark}{\ding{51}}%
\newcommand{\xmark}{\ding{55}}%

\newcommand{\fasttext}{{\sc FastText}}
\newcommand{\wordtovec}{\textsc{Word2Vec}}
\newcommand{\glove}{\textsc{GloVe}}

\title{Emotion Representation Mapping for Automatic Lexicon Construction (Mostly) Performs on Human Level}

\author{Sven Buechel \& Udo Hahn\\
\texttt{\{sven.buechel|udo.hahn\}@uni-jena.de}\\ 
Jena University Language \& Information Engineering (JULIE) Lab\\
Friedrich-Schiller-Universit\"at Jena, Jena, Germany\\
\url{http://www.julielab.de}}

\date{}

\begin{document}
\maketitle
\begin{abstract}
Emotion Representation Mapping (\thistask{}) has the goal to convert existing emotion ratings from one representation format into another one, e.g., mapping Valence-Arousal-Dominance annotations for words or sentences into Ekman's Basic Emotions and vice versa. \thistask{} can thus not only be considered as an alternative to Word Emotion Induction (WEI) techniques for automatic emotion lexicon construction but may also help mitigate problems that come from the proliferation of emotion representation formats in recent years. We propose a new neural network approach to \thistask{} that not only outperforms the previous state-of-the-art. Equally important, we present a refined evaluation methodology and gather strong evidence that our model yields results which are (almost) as reliable as human annotations, even in cross-lingual settings. Based on these results we generate new emotion ratings for 13 typologically diverse languages and claim that they have near-gold quality, at least.
\end{abstract}

%
%
\blfootnote{
    %
    %
    %
    %
    %
    %
    \hspace{-0.65cm}  
    This work is licensed under a Creative Commons 
    Attribution 4.0 International License.
    License details:
    \url{http://creativecommons.org/licenses/by/4.0/}
}

\input{intro}

\input{related}

\input{methods}

\input{results}
\input{conclusion}

\section*{Acknowledgements}
We thank the anonymous reviewers for their thoughtful comments and suggestions.

\bibliographystyle{stylesheet/acl}
\bibliography{literature_SB-coling18}

\end{document}

%% file: intro.tex
\section{Introduction}
\label{sec:intro}

From its inception, researchers in the field of sentiment analysis aimed at predicting the affective state that is typically associated with a given word based on a list of linguistic features, a problem referred to as \textit{word emotion induction} (WEI)  \cite{Hatzivassiloglou97acl,Turney03}.
Early research activities have focused on {\it semantic polarity} (the positiveness or negativeness of a feeling) for quite a long time. But more recently this focus on binary representations has been replaced by more expressive \textit{emotion representation formats} such as Basic Emotions or Valence-Arousal-Dominance. In the meantime, WEI has become an active area of research, regularly featured in shared tasks \cite{Rosenthal15,Yu16ialp}. 
Based on these achievements, WEI techniques have become a natural methodological choice for the automatic construction of emotion lexicons \cite{Koeper16,Shaikh16}.

Yet, only very recently, a radically different approach to automatic emotion lexicon construction has been proposed. Instead of relying on linguistic features (such as similarity with seed words or word embeddings), the goal of \textit{emotion representation mapping} (\thistask) is to  derive new emotional word ratings \textit{in one format} based on known ratings of the same words \textit{in another format} \cite{Buechel17cogsci}. For example, \thistask{} could use empirically gathered ratings for Basic Emotions and convert them into a Valence-Arousal-Dominance representation scheme, with greater precision than currently achievable by  WEI algorithms. As a much appreciated side effect, one of the promises of \thistask{} is to make otherwise incompatible resources (lexicons or annotated corpora, as well as tools) compatible, and incomparable systems comparable. Thus, this approach has the potential to mitigate some of the negative effects that arise from not having a community-wide standard for emotion annotation and representation \cite{Calvo13,Buechel18lrec}.

We here want to contribute to this endeavor by providing a large-scale evaluation of previously proposed \thistask{} approaches for four typologically diverse languages and  report evidence that \thistask{} clearly outperforms current state-of-the-art WEI algorithms. Furthermore, we present our own deep learning model which performs even better against all competitors. Most importantly, however, we propose a new methodology for comparing the reliability of \thistask{} against human annotation reliability, a major shortcoming of previous work. 
As a result, we find that our proposed model performs competitive to a reasonably large group of human raters, \textit{even in cross-lingual settings}. 
Based on this evidence, we automatically construct emotion lexicons for 13 languages and claim that they have (near) gold quality. These lexicons as well as our experimental code base and results are publically available.\footnote{\label{fn:github}\url{https://github.com/JULIELab/EmoMap}}

%% file: related.tex
\section{Related Work}
\label{sec:related}

\paragraph{Psychological Models of Emotion.}

\input{figs/vadcube.tex}

Models of emotion typically fall into two main groups, namely  {\it discrete} (or {\it categorical}) and {\it dimensional} ones \cite{Stevenson07,Calvo13}. Discrete models are built around particular sets of emotional categories deemed fundamental and universal. \newcite{Ekman92}, for instance, identifies six {\it Basic Emotions} (Joy, Anger, Sadness, Fear, Disgust and Surprise). 
In contrast, dimensional models consider emotions to be composed out of several influencing factors (mainly two or three). These are often referred to as {\it Valence} (corresponding to the concept of polarity), {\it Arousal} (a calm--excited scale), and {\it Dominance} (perceived degree of control over a (social) situation)---the VAD model. 
The last dimension, Dominance, is quite often omitted, thus constituting the VA model. For convenience, both will be jointly referred to as VA(D).
An illustration of VAD and its relationship to Basic Emotions is given in Figure \ref{fig:vadcube}.

\paragraph{Lexical Data Sets.}

In contradistinction to NLP where many different representation formats for emotions are being used, lexical resources originating from psychology labs almost exclusively subscribe either to VA(D) or Basic Emotions models (typically omitting Surprise; the BE5 format). 
Over the years, a considerable number of resources built on these premises have emerged from psychological research for various languages.\footnote{See, e.g., Tables \ref{tab:data} and \ref{tab:constructed}. An enhanced list of these and similar data sets is provided in \newcite{Buechel18lrec}.}
In more detail, these lexical ratings have been gathered via  questionnaire studies by collecting individual ratings from a large number of subjects for each lexical item under consideration (typically between 20 to 30 individual ratings per item). These individual assessments are then averaged to yield aggregated scores on which we base our experiments. The emotion values we deal with must thus be understood as an average emotional reaction when presenting a lexical stimulus to a group of human judges.

\input{tabs/data}

In this paper, we restrict ourselves to the VA(D) and BE5 format. Following the conventions of the emotion lexicons used in our experiments (Table \ref{tab:data}),
each VA(D) dimension receives a value from the interval $[1,9]$ where `1' means ``most negative/calm/submissive'', `9' means ``most positive/excited/dominant'' and `5' means ``neutral''. Conversely, values for  BE5 categories range in the interval $[1,5]$ where `1' means ``absence'' and `5' means ``most extreme'' expression of the respective emotion.\footnote{
	Although these intervals are fairly well established conventions, in some data sets different rating scales were used, nevertheless. In these cases, we linearly transformed the ratings so that they match the defined intervals.
} 
Consequently, the VA(D) and BE5 formats are conceptually different from one another insofar as VA(D) dimensions are bi-polar, whereas BE5 categories are uni-polar.

\paragraph{Word Emotion Induction.}

Automatically constructing such word-level emotion data sets has been a focus of NLP-based sentiment analysis studies from the beginning. In fact, the problem to automatically predict polarity or emotion scores for a given word based on some linguistic features---often referred to as Word Emotion Induction (WEI)---is already dealt with in the seminal work of  \newcite{Hatzivassiloglou97acl}.  At first, the features taken into account were typically derived from co-occurrence or terminology-based similarity  with a small set of \textit{seed word} with known emotional scores \cite{Turney03,Esuli05cikm}. Nowadays, these features are almost completely replaced by \textit{word embeddings}, i.e.,
dense, low-dimensional vector representations of words that are trained on large volumes of raw text in an unsupervised manner.  \wordtovec{} \cite{Mikolov13nips}, \glove{} \cite{Pennington14} and \fasttext{} \cite{Bojanowski17tacl} are among today's most popular algorithms for generating embeddings.

WEI algorithms constitute a natural baseline  for \thistask{} because, first, they produce the same output (emotion ratings for words according to some emotion representation format), yet their predictions are based on expressively weaker features (word embeddings instead of emotion ratings for the same word but in another format), thus constituting a harder task. Second, they form the currently prevailing paradigm for the automatic construction of emotion lexicons  \cite{Koeper16,Shaikh16}, a problem for which \thistask{} offers a promising alternative.

\paragraph{Emotion Representation Mapping.}
\label{sec:related.mapping}

\input{tabs/example_entries}

In contrast to WEI, \thistask{} is based on the condition that the pairs of data sets in Table \ref{tab:data} are complementary in the sense that, when combining these lexicons, a subset of their entries are then encoded in {\it both} emotion formats, i.e., VA(D) {\it and} BE5.
This condition is illustrated for three lexical items in Table \ref{tab:examples}.

Although such complementary data sets have been available for quite some time, \thistask{} has only recently been introduced to NLP by \newcite{Buechel16ecai} in order to compare a newly proposed VAD-based prediction system against previously established results on Basic Emotion gold standards.
In a follow-up study, \newcite{Buechel17eacl} devised \textsc{EmoBank}, a VAD-annotated corpus which, in part, also bears BE5 ratings on the \textit{sentence} level. 
They found that both kinds of annotation were highly predictive for each other using a $k$-Nearest-Neighbor approach. 
In later studies, they examined the potential of \thistask{} as a substitute for manual annotation of \textit{lexical} items, also in cross-lingual settings \cite{Buechel17cogsci,Buechel18lrec}. Although their evaluation was limited in expressiveness, they already found evidence that \thistask{} may be comparable to human performance in terms of the quality of the resulting ratings.

Similar work has, to the best of our knowledge, only been done in the psychology domain. 
However, related work from this area does not target the goal of predictive modeling \cite{Stevenson07,Pinheiro17}. 
In both contributions, linear regression models were fitted to predict VAD dimensions given BE5 categories and vice versa. 
Yet, this was mainly done to inspect the respective slope-coefficients as an indicator of the relationship of dimensions and categories. 
Thus, the overall goodness of the fit was \textit{not} in the center of interest and was not even reported by \newcite{Stevenson07}.

%% file: figs/vadcube.tex
\begin{wrapfigure}{R}{.5\textwidth}
	\centering
	\includegraphics[width=.5\textwidth]{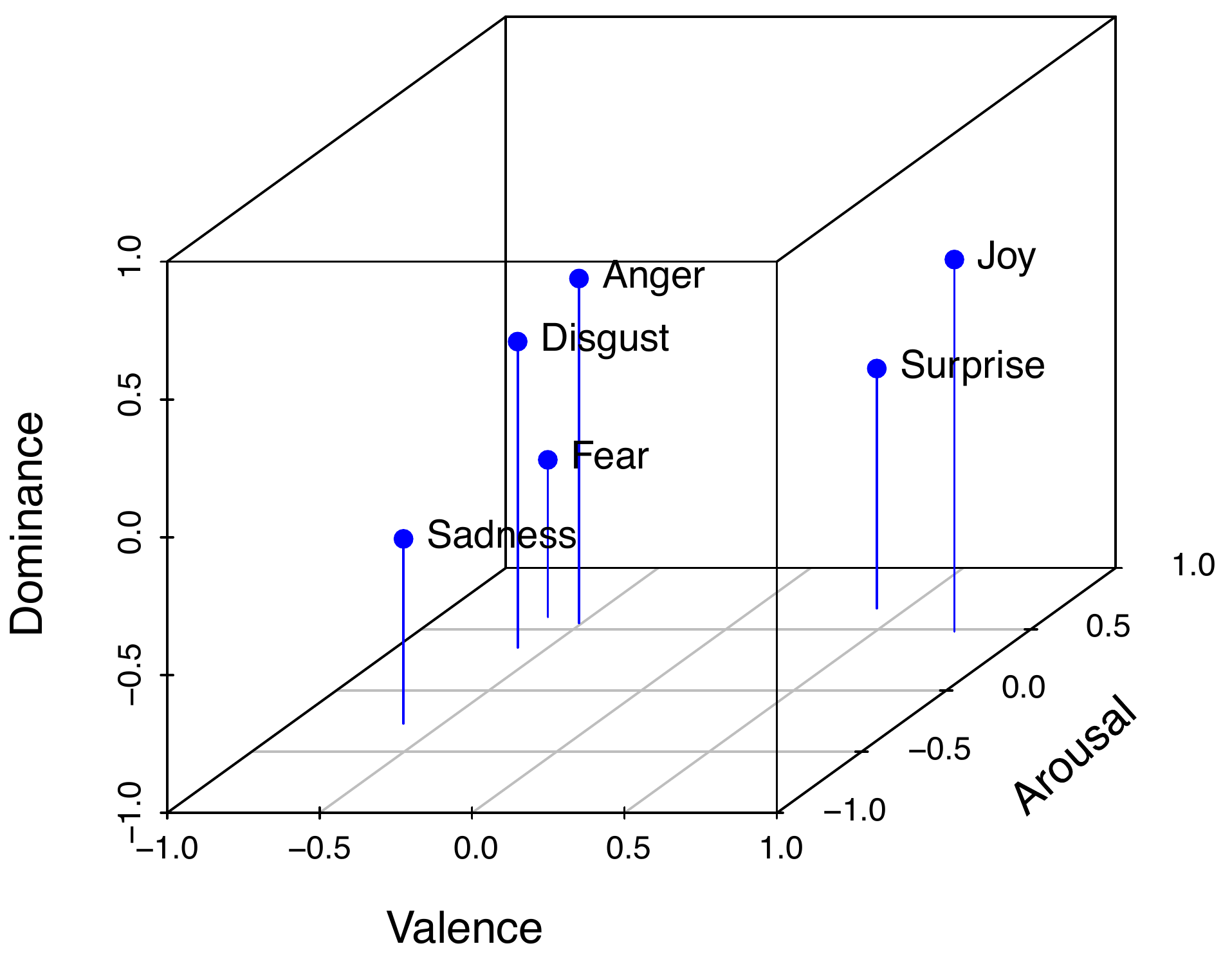}
	\caption{ \label{fig:vadcube}
	Affective space spanned by the Valence-Arousal-Dominance (VAD) model, together with the position of six Basic Emotions. Adapted from \newcite{Buechel16ecai}.} 
\end{wrapfigure}

%% file: tabs/data.tex
\begin{table}[h!]
	\centering
	\small
	\begin{tabular}{|lllcr|}
		\hline
		Abbrev. & VA(D)  & BE5 & Dom? & Overlap\\
		\hline\hline
		en\_1 & \newcite{Bradley99anew}  & \newcite{Stevenson07} & \cmark{} & 1,028\\
		en\_2 & \newcite{Warriner13}  & \newcite{Stevenson07}& \cmark{} & 1,027\\
		es\_1 & \newcite{Redondo07}  & \newcite{Ferre17} & \cmark{}& 1,012\\
		es\_2 & \newcite{Hinojosa16dom}  & \newcite{Hinojosa16}& \cmark{} & 875\\
		es\_3 & \newcite{Stadthagen16}  & \newcite{Stadthagen17}& \xmark{} & 10,491\\
		de\_1 & \newcite{Vo09}  & \newcite{Briesemeister11} & \xmark{}& 1,958 \\
		pl\_1 & \newcite{Riegel15}  & \newcite{Wierzba15}& \xmark{} & 2,902\\
		pl\_2 & \newcite{Imbir16}  & \newcite{Wierzba15} & \cmark{} & 1,272\\
		\hline
	\end{tabular}
	\caption{\label{tab:data}
		Data sets used in our experiments; with abbreviation (including language code according to ISO 639-1), the bibliographic sources of the VA(D) and BE5 ratings, information on whether Dominance is included and the number of overlapping entries.}
\end{table}

%% file: tabs/example_entries.tex
\begin{wraptable}{r}{8.2cm}
	\small
	\centering
	\begin{tabular}{|l|ccc|ccccc|}
		\hline 
		Word & V & A & D & J & A & S & F & D\\
		\hline \hline
		\textit{sunshine} 	& 8.1 & 5.3 & 5.4 & 4.3 & 1.2 & 1.3 & 1.3 & 1.2\\
		\textit{terrorism}	& 1.6 & 7.4 & 2.7 & 1.1 & 3.0 & 3.4 & 4.1 & 2.5\\
		\textit{orgasm} 	& 8.0 & 7.2 & 5.8 & 4.3 & 1.3 & 1.3 & 1.4 & 1.2\\
		\hline
	\end{tabular}
	\caption{\label{tab:examples}
		Three lexical items and their emotion values in VAD (second column group) and BE5 (third column group) format. VAD scores are taken from \protect\newcite{Warriner13}, BE5 scores were automatically derived (see Section \ref{sec:results.lexicon_construction}).
		}
\end{wraptable}

%% file: methods.tex
\section{Methods}
\label{sec:methods}

Let $L \coloneqq \{w_1, w_2, ..., w_n\}$ be a set of words. 
Let $s$,$t$ denote two distinct \textit{emotion representation formats} such that \textit{both} $emo^s(w_i) \in \mathbb{R}^{|s|}$ and $emo^t(w_i) \in \mathbb{R}^{|t|}$ describe the emotion vector associated with $w_i$ relative to $s$ and $t$, respectively, where $|s|$, $|t|$ denote the number of variables which each format employs (e.g.,  3 for VAD and 5 for BE5).
The task we address in this paper is to predict the \textit{target emotion ratings} $T \coloneqq \{ emo^t(w_i)|\; w_i \in L\}$ given the set $L$ and the corresponding \textit{source emotion ratings} $S \coloneqq \{ emo^s(w_i)|\; w_i \in L\}$.
Performance will be measured as Pearson correlation $r$ between the predicted values and human gold ratings (one $r$-value per element of the target representation). 
In general, the Pearson correlation between two data series $X:=x_1, x_2, ..., x_n$ and $Y:=y_1, y_2, ..., y_n$ takes values between $+1$ (perfect positive correlation) and $-1$ (perfect negative correlation) and is computed as 
\begin{equation}
r_{xy} \coloneqq \frac{\sum_{i=1}^n (x_i-\bar{x})(y_i-\bar{y})}{\sqrt{\sum_{i=1}^n(x_i-\bar{x})^2} \; \sqrt{\sum_{i=1}^n(y_i-\bar{y})^2}}
\end{equation}
where $\bar{x}$ and $\bar{y}$ denote the mean values for $X$ and $Y$, respectively.

\subsection{Reference Methods}
\label{sec:methods.reference}

The first method against which we will compare our proposed model is linear regression (LR) as used by \newcite{Stevenson07} in their early study. LR predicts an emotion value in the target representation $t$ as the affine transformation
\begin{equation}
emo_\text{LR}^t(w_i) \coloneqq W\; emo^s(w_i) +b
\end{equation}
where $W$ is a $|t| \times |s|$ matrix and $b$ is a $|t| \times 1$ vector. The model parameters are fitted using ordinary least squares.
In contrast, \newcite{Buechel17eacl} proposed the use of \textit{k}-Nearest-Neighbor Regression (KNN) for \thistask{}. This simple supervised approach predicts the target value as
\begin{equation}
emo^t_\text{KNN}(w_i) \coloneqq \frac{1}{k} \sum_{{w_i}' \in \text{NEAREST}(w_i,k,S)} emo^t({w_i}')
\end{equation}
where NEAREST yields the $k$ nearest neighbors of $w_i$ in the training set (determined by the Euclidean distance between the source representations of two words).  The $k$ parameter was fixed to $20$ based on a pilot study.\footnote{
	In contrast, \newcite{Buechel17cogsci} determined $k$ for each lexicon \textit{individually} based on a dev set. Now, we deviate from this approach  since it is inapplicable for the cross-lingual lexicon construction presented in Section \ref{sec:results.lexicon_construction}.
} 
We used the \url{scikit-learn.org} implementation for both LR and KNN.

\subsection{Proposed Model: A Multi-Task Feed-Forward Neural Network for \thistask{}}
\label{sec:methods.model}

Despite the fact that the above set-ups already perform quite well for \thistask{} (see Section \ref{sec:results}), both LR and KNN are rather basic types of models lacking deeper sophistication.  
As a consequence, we here propose the use of Feed-Forward Neural Networks\footnote{
	Note that applying neural architectures currently  popular for other NLP tasks is not advisable because of the simplicity of our input data (feature vectors of length 2 to 5). These more complex architectures are instead designed for, e.g., \textit{sequential} data (such as the RNN family) or \textit{spatially arranged} data (such as CNNs).
} 
(FFNNs) for \thistask{} which have been shown to be capable of approximating arbitrary functions, in theory at least \cite{Hornik91}.
In general, an FFNN consists of an {\it input layer} with activation $a^{(0)} \coloneqq emo^s(w_i) \in \mathbb{R}^{|s|}$ followed by  multiple hidden layers with activation $a^{(l+1)} \coloneqq \sigma(W^{(l+1)}a^{(l)}+b^{(l+1)})$ where $W^{(l+1)}$ and  $b^{(l+1)}$ are the weights and biases for layer $l+1$ and $\sigma$ is a nonlinear activation function.
Since the emotion formats under scrutiny capture affective states as real-valued vectors, the activation on the output layer $a^{out}$ (where $out$ is the number of non-input layers in the network) is computed as the affine transformation
\begin{equation}\label{eq:model}
emo^t_\text{FFNN}(w_i) \coloneqq a^{(out)} \coloneqq W^{(out)}a^{(out-1)}+b^{(out)}
\end{equation}

Consequently, our model differs from the other approaches presented in this section by \textit{sharing} model parameters (weights and biases of the hidden layers) across  the different dimensions/categories of the target format with only the last layer having parameters which are uniquely associated to one of the outputs (see Equation \ref{eq:model}).
This can be considered as a mild form of multi-task learning \cite{Caruana97}, a machine learning technique which has been shown to strongly decrease the risk of overfitting \cite{Baxter97} and also speeds up computation by greatly decreasing the number of tunable parameters compared to training individual layers for each affective dimension/category.

The remaining specifications of our model are as follows. We train two-hidden layer FFNNs (both with 128 units), ReLU activation, $.2$ dropout on the hidden layers (none on the input layer)\footnote{We found the usual recommendation of $.2$ on input and $.5$ on hidden layers \cite{Srivastava14} too high given the small number of features in our task (2 to 5).} and Mean-Squared-Error loss. Each model was trained for $10,000$ iterations (well beyond convergence, independently of the size of the training set) using the \textsc{Adam}  optimizer \cite{Kingma15}. \mbox{\url{Keras.io}} was used for implementation.

\subsection{Baseline: Word Emotion Induction}
\label{sec:methods.baseline}

As a natural baseline for \thistask{}, we will use a recent state-of-the-art method for word emotion induction (WEI) by \newcite{Du16}.\footnote{
	In our most recent contribution featuring a large-scale evaluation of many current WEI approaches on numerous data sets, we found that among the existing ones the model proposed by \newcite{Du16} performs best, only beaten by our own, newly proposed model \cite{Buechel18naacl}. Note that even compared to this more advanced approach to WEI, the performance figures we report here for \thistask{} still remain much higher (see Section \ref{sec:results}). Hence, the claim of this paper that ERM is superior to WEI, remains valid even despite most recent achievements for the latter task.
}
They propose Feed-Forward Neural Networks (similar to our proposed model for \thistask{}) in combination with a boosting algorithm.
The authors used FFNNs with a single hidden layer of $100$ units and ReLU activation. The boosting algorithm \textsc{AdaBoost.R2} \cite{Drucker97}  was used to train the ensemble (one per target variable). 
We implemented this approach with \texttt{scikit-learn} using exactly  the same settings as in the original publication.\footnote{
	Publicly available at: \url{https://github.com/StevenLOL/ialp2016_Shared_Task}
} 
As for the word embeddings this method needs as input, we used the pre-trained \fasttext{} embeddings that Facebook Research makes available for a wide range of languages trained on the respective Wikipedias.\footnote{\url{https://github.com/facebookresearch/fastText/blob/master/pretrained-vectors.md}} This way, we hope to achieve a particularly high level of comparability across languages because, for each of them, embeddings are trained on data from the same domain and of a similar order of magnitude.\footnote{
	For English, much larger embedding models are publicly available, yet not for the other languages under consideration; cf. \newcite{Buechel18naacl}.
}

\subsection{Comparison to Human Reliability}
\label{sec:reliabilities}

Since common metrics for Inter-Annotator Agreement (IAA), such as Cohen's Kappa, are not applicable for real-valued emotion scores \cite{Carletta96}, we will now  discuss how to compare our own results against human assessments in order to put their reliability on a safe ground.

One possible point of comparison that has been used in previous work \cite{Buechel17cogsci,Buechel18lrec} is \textit{inter-study reliability} (ISR), i.e., the correlation between the ratings of common words in different data sets. However, this procedure comes with a number of downsides. First, the number of pairs of data sets with substantially overlapping entries is rather small since researchers focus mainly on acquiring ratings for \textit{novel} words instead of gathering annotations anew for ones already covered. Thus, employing ISR comparison with human performance is only possible on few data sets. In particular, we are not aware of any pair of data sets with significantly overlapping BE5 ratings. Second, ISR is sensitive to differences in acquisition methodologies (e.g., alternative sets of instructions or rating scales) and may thus vary substantially between different pairs of data sets.

As an alternative, these shortcomings lead us to propose \textit{split-half reliability} (SHR) as a new basis for our comparison. SHR is computed by splitting all individual ratings for each of the items into two groups. These individual ratings are then averaged for both groups and the Pearson correlation between the group averages is computed. The whole processes is repeated (typically 100 times) with random splits before averaging the results from each iteration \cite{Mohammad17starsem}. Thus, an important difference between SHR and ISR is that the former is computed on a \textit{single} data set whereas the latter requires two \textit{different} data sets with overlapping items. On the other hand, ISR can be computed on the final ratings alone, whereas SHR requires knowledge of the judgments of the individual raters. Most often, these individual ratings are not distributed. Yet, luckily, SHR values are commonly reported when publishing emotion lexicons (see below).

Still, both SHR and ISR---as well as other popular approaches to reliability estimation for numerical emotion scores, e.g., the leave-one-out approach presented by \newcite{Strapparava07}---are heavily influenced by the number of participants of a study. For SHR, this is intuitively clear because with enough subjects, both groups should yield reliable estimates of the true population mean ratings, leading to very high correlation values between the groups. 
As a result, by splitting the number of raters into two groups for the SHR estimate, this technique will on average produce lower correlation values than if the study was repeated with the full number of participants and correlation between the first and second study had been computed (test-retest reliability). 
To counterbalance this effect, when reporting SHR values, authors often turn to  \textit{Spearman-Brown adjustment} (SBA; \newcite{Vet17}), a technique which estimates the reliability $r^{*}$ of a study if the number of subjects was increased by the factor $k$:
\begin{equation}\label{eq:sba}
r^{*} \coloneqq \frac{k\;r}{1+(k-1)\;r}
\end{equation}
were $r$ is the \textit{empirically measured} SHR and $k$ is set to $2$ for the use case discussed above (virtually doubling the number participants). 

Since some authors of the data sets in Table \ref{tab:data} apply SBA while others do not, the reported SHR values must be normalized to guarantee a consistent evaluation. 
Going one step further, we can even apply SBA to normalize the reported values with respect to the number of participants in a given study, thus establishing an even more consistent ground for evaluation.

\input{tabs/reliabilities}

We chose the \textit{normalized number of participants} to be 20, i.e., the adjusted scores (reported in Table \ref{tab:reliabilities}) estimate the \textit{empirical} SHR values, if the given study was conducted with 20 participants (the average correlation between two randomly assigned groups of 10 raters). Normalization was conducted by applying Equation (\ref{eq:sba}) to the reported values with $k:=N^*/N$, if SBA was not already applied, or $k:=N^*/(2 \times N)$, if SBA was already applied to the reported values; $N$ being the actual number of participants and $N^* := 20$ being the normalized number of participants.

It is important to note that the decision for $N^{*} = 20$ is necessarily arbitrary, to some degree, with higher SHR estimates arising from higher values of $N^{*}$. However, 20 raters are often used in psychological studies \cite{Warriner13,Stadthagen16}, while being way higher than the number of raters typically used in NLP for emotion annotation,  both for the word and sentence level \cite{Yu16naacl,Strapparava07}. Thus, we argue that this choice constitutes a rather challenging line of comparison for our system.

Since model performance will be measured in terms of Pearson correlation (see above), the performance figures achieved on the gold data can be compared with the adjusted SHR (also based on correlation). We can interpret cases where the former outperforms the latter as \textit{the model agreeing more with the gold data than two random groups of ten annotators would agree with each other}. Thus, for these cases we say our model achieves \textit{super-human}  performance, as it cannot be expected that a well-conducted annotation study leads to more reliable results.

%% file: tabs/reliabilities.tex
\begin{wraptable}{R}{9cm}
\centering
\small
\begin{tabular}{|l|rrr|rrrrr|}
\hline
{} &  Val &  Aro &  Dom &  Joy &  Ang &  Sad &  Fea &  Dsg \\
\hline\hline
en\_1 &  --- &  --- &  --- &  --- &  --- &  --- &  --- &  --- \\
en\_2 & .914 & .689 & .770 &  --- &  --- &  --- &  --- &  --- \\
es\_1 &  --- &  --- &  --- & .915 & .889 & .915 & .889 & .864 \\
es\_2 & .839 & .730 & .730 & .915 & .915 & .915 & .889 & .889 \\
es\_3 & .880 & .750 &  --- & .754 & .786 & .818 & .802 & .739 \\
de\_1 &  --- &  --- &  --- &  --- &  --- &  --- &  --- &  --- \\
pl\_1 & .928 & .630 &  --- & .884 & .802 & .821 & .821 & .802 \\
pl\_2 & .935 & .679 & .725 & .884 & .802 & .821 & .821 & .802 \\
\hline
\end{tabular}
\caption{\label{tab:reliabilities}
	Normalized split-half reliabilities for VAD and BE5 for the data sets used in our experiments. ``---'' indicates that reliability has not been reported.}
\end{wraptable}

%% file: results.tex
\section{Results}
\label{sec:results}

\subsection{Ablation Experiments on Affective Dimensions and Categories}
\input{figs/ablation}

Previous work has limited itself to data sets comprising all three VAD dimensions with the implicit belief that Dominance provides valuable affective information which is important for \thistask{}. However, since only about half of the data sets developed in psychology labs (and even less provided by NLP groups) actually \textit{do} comprise Dominance, this decision massively decreases the amount of data sets at hand. 
To resolve this dilemma, the following experiment aims at quantifying the relative importance of the different affective variables of the VAD and the BE5 format.

Our set-up works as follows: For each data set from Table \ref{tab:data} that includes the Dominance dimension, we trained one LR model\footnote{
	Linear regression was used because it does not comprise any hyperparameters that might heavily influence the outcome of this experiment (thus leading to greater generality of the results).
} 
(Section \ref{sec:methods.reference}) to map VAD to BE5 and another one to map BE5 to  VAD (`dim2cat' and `cat2dim'  for short) applying 10-fold cross-validation. The resulting performance measurements were averaged over all data sets.

We then repeated this procedure once for each VAD dimension (when mapping dim2cat) and each BE5 category (when mapping cat2dim), omitting  one of the dimensions/categories from the source representation in every iteration, thus constituting a kind of ablation experiment.
Next, for each of the ``incomplete'' models, we computed the difference between its performance and the performance of the ``complete'' model (not lacking any of the variables). 
Now, we can use this loss of performance as an estimate of the \textit{relative importance} of the respective left-out dimension or category. The results of this experiment are depicted in Figure \ref{fig:ablation}.

As can be seen, regarding VAD, Valence is by far the most important dimension with a performance drop of .12 when ablating it. In turn, Arousal, the second-best  dimension only increases performance by .04, whereas Dominance contributes to less than .01 of the performance.
Similarly, for Basic Emotions, Joy is the most important category, although BE5 seems to distribute the affective information more equally across its variables (with the exception of Disgust which contributes  far less than .01 to the performance).

Since our data suggest that Dominance  plays only a minor role within the VAD framework, we will \textit{not} limit our further experiments to data sets including this dimension---as it was done in previous work (Section \ref{sec:related.mapping})---but rather include the large variety of bi-representational data sets which leave it out (see Table \ref{tab:data}).

\subsection{Monolingual Representation Mapping}
\label{sec:results.monolingual}

\input{tabs/monolingual}

In this experiment, we compared the performance of the WEI baseline, the LR- and KNN-based reference methods for \thistask{} and our newly proposed FFNN model. For each of these methods and data sets in Table \ref{tab:data}, we trained one model to map cat2dim and another one to map dim2cat (for the \thistask{} methods) or to predict VA(D) ratings and BE5 ratings based on word embeddings 
for the WEI baseline. The whole process was conducted using 10-fold cross-validation where we used identical train/test splits for all methods.\footnote
	{This procedure constitutes a more direct comparison than using different splits for each method and allows {\it paired} $t$-tests.
	} 
The results of this experiment are displayed in Table \ref{tab:monolingual_average}, only showing the average values over VA(D) and BE5, respectively, but allowing for an easy comparison between the different approaches.

As can be seen, all of the \thistask{}  approaches (LR, KNN, FFNN) perform  more than 10\%-points better than the state of the art in word emotion induction (WEI) for VAD prediction and at least about 5\%-points better for BE5 predictions (on average over all data sets and affective variables). This finding already strongly suggests that \thistask{} is the superior approach for automatic lexicon creation, given that the required data are available. This might be especially useful in situations where, say, large VAD  but only small BE5 lexicons are available for a given language (see Section \ref{sec:results.lexicon_construction}).
Regarding the ordering of the \thistask{} approaches, KNN outperforms LR in almost all cases. The advantage is more pronounced for mapping dim2cat (2.5\%-points difference on average) than cat2dim (.4\%-points difference).
On top of that, our proposed FFNN model outperforms KNN by a 1.2\%-point margin for cat2dim and a .8\%-point margin for dim2cat (again as average over all data sets) performing best on each single data set. Regarding the 16 cases of Table \ref{tab:monolingual_average} (8 data sets times two mapping directions), the performance gain of FFNN compared to the respective second best system is statistically significant\footnote{
	Paired two-tailed $t$-tests based on the 10 train/test splits during cross-validation; $p<.05$.
} in all but 2 cases.
The differences between the individual \thistask{} approaches might appear quite small, yet become a lot more meaningful considering the proximity to human annotation capabilities as discussed in the following paragraphs.

Table \ref{tab:monolingual_individual} displays the performance figures of the FFNN model relative to each affective variable. As can be seen, among VAD, Valence is the easiest dimension to predict ($r=.956$ on average over all data sets) whereas for Arousal the performance is worst . Similarly, for BE5, Joy obtains the best values ($r=.932$) and Disgust is the hardest to predict. Interestingly, the overall ordering of performance within the two formats is consistent with the ordering of human reliability (see Table \ref{tab:reliabilities}).

Comparing our system performance against human SHR (based on 20 participants per study; see Section \ref{sec:reliabilities}), again our approach seems to be highly reliable (color coding of Table \ref{tab:monolingual_individual}). In particular, \thistask{} using the FFNN model outperforms SHR in over half of the applicable cases (25 of 38). For mapping cat2dim it surpasses human reliability in all but 2 cases whereas when mapping dim2cat the reported SHR is surpassed in over half of the cases (14 out of 25). 

This result, astonishing as it might appear, is yet consistent with findings from previous work which, in turn, were based on ISR (not on SHR) data \cite{Buechel17cogsci,Buechel18lrec}. We conclude that in the monolingual set-up, \thistask{} using the FFNN model substantially outperforms current capacities in word emotion induction and is even more reliable than a medium sized human rating study. Thus these automatically produced ratings should be cautiously attributed gold standard quality.

\subsection{Crosslingual Representation Mapping} 
\label{sec:results.crosslingual}

\input{tabs/multilingual}

In the crosslingual  set-up, we make use of the fact that our model does not rely on any language-specific information, since the categories/dimensions describe supposedly universal affective states rather than linguistic entities.
Thus, models trained on one language could, in theory, be applied to another one without any need for adaptation. This capability comes in handy when only data sets according to \textit{one} emotion format exist for a given language. In such cases we could still train our model on data available for other languages and use it to produce new ratings for the language in focus. This section aims at estimating the performance of lexicons derived in this manner.

For each of the data sets in Table \ref{tab:data}, we trained FFNN models to map cat2dim and dim2cat, respectively. We trained on each gold lexicon that did not cover the language of the data set under scrutiny (e.g., for testing on en\_1, the models were trained on all Spanish, Polish and German data sets, but not on en\_2). Since this set-up leads to fixed train and test sets, we did not perform cross-validation. For comparability between data sets, the Dominance dimension was excluded for this experiment.

Overall, the results remained astonishingly stable compared to the monolingual set-up, with performance figures for Valence and Joy dropping by less than 1\%-point on average over all data sets (see Table \ref{tab:crosslingual}). Also, Anger, Sadness, Fear and Disgust only suffer a moderate decrease of about 5\%-points at most---only the performance of Arousal decreased more than that. 

A possible explanation for these strong results is the marked increase in the amount of training data that comes along with training on the majority of the available data (independent of language). This circumstance seems to counterbalance much of the negative effects that may arise in this crosslingual applications.

In comparison to SHR, the \thistask{} approach  still turns out to work quite well. Regarding VA, we outperform human reliability in 8 of 10 cases. Concerning BE5, SHR was beaten in about half of the cases (11 of 25). We conclude that, although the capability of our mapping approach suffers a bit in the crosslingual set-up, it still produces very accurate predictions and can thus be attested \textit{near} gold quality, at least.

\subsection{Automatic Lexicon Construction for Diverse Languages}
\label{sec:results.lexicon_construction}

\input{tabs/constructed}

After the positive evaluation of the FFNN model for \thistask{}, the last bit of our contributions is to apply the created models to a wide variety of data sets which so far  bear emotion ratings for {\it one} format only (either VA(D) or BE5). Based on the experiments reported so far, we claim that these have gold quality (for the monolingual approach, Section \ref{sec:results.monolingual}) or near-gold quality (for the crosslingual approach, Section \ref{sec:results.crosslingual}).

For the monolingual approach, we train our model on the data set on which we achieved the highest performance in Section \ref{sec:results.monolingual} for the respective language (assuming this hints at particularly ``clean'' data). 
In contrast, in the crosslingual set-up, training data are acquired by concatenating \textit{all} the available data sets from Table \ref{tab:data} (consequently ignoring Dominance for compatibility).

Table \ref{tab:constructed} lists the emotion lexicons constructed in this manner together with their most important characteristics. 
The number of new ratings ranges from almost 13,000 (for English) and 10,500 (for Spanish), over several thousands (for Dutch, Chinese and Polish, ) and around 1,500--1,000 (for Indonesian, Italian, Portuguese, Greek, French and  German) to 200--100 (for Finnish and Swedish). 
For illustration, Table \ref{tab:examples} displays three entries of the English BE5 lexicon, the largest one we constructed.

%% file: figs/ablation.tex
\begin{wrapfigure}{R}{0.45\textwidth}	
	\centering
	\includegraphics[width=.45\textwidth]{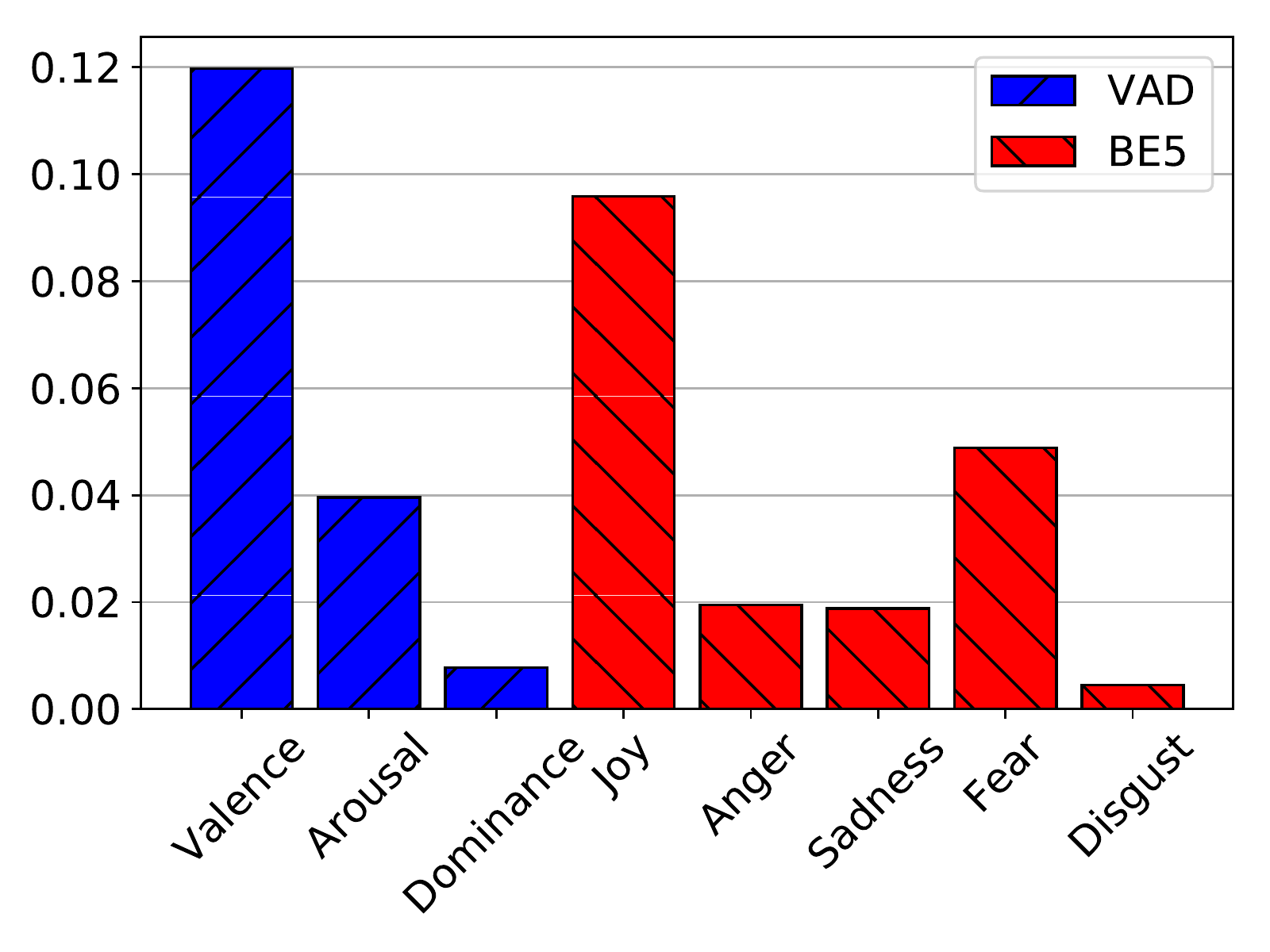}
	\caption{\label{fig:ablation}
		Relative importance of the affective variables of VAD and BE5 for predicting the alternative format, respectively; measured in drop of Pearson $r$ when using all variables vs.\ omitting the one under scrutiny.}
\end{wrapfigure}

%% file: tabs/monolingual.tex
\begin{wraptable}{R}{10cm}
\captionsetup[sub]{font=normalsize,labelfont=normalsize}
\centering
\begin{subtable}{10cm}
\centering
\small
\begin{tabular}{|p{.5cm}|rrrr|rrrr|}
\hline
{} & \multicolumn{4}{c|}{cat2dim} & \multicolumn{4}{c|}{dim2cat} \\
{} & WEI & LR & KNN & FFNN & WEI & LR & KNN & FFNN \\
\hline\hline
en\_1    &            .685 &                \underline{.841} &                 .840 &            {\bf.853}**\phantom{*} &            .818 &                .844 &                 \underline{.868} &            {\bf.877}*\phantom{**} \\
en\_2    &            .741 &                .827 &                 \underline{.828} &            {\bf.843}*** &            .821 &                .829 &                 \underline{.852} &            {\bf.858}*** \\
es\_1    &            .709 &                \underline{.856} &                 .855 &            {\bf.869}*** &            .775 &                .804 &                 \underline{.849} &            {\bf.853}\phantom{***} \\
es\_2    &            .600 &                .823 &                 \underline{.828} &            {\bf.844}*** &            .797 &                .863 &                 \underline{.882} &            {\bf.889}*\phantom{**} \\
es\_3    &            .713 &                \underline{.799} &                 .796 &            {\bf.804}*** &            .743 &                .776 &                 \underline{.820} &            {\bf.826}*** \\
de\_1    &            .758 &                .819 &                 \underline{.827} &            {\bf.837}**\phantom{*} &            \underline{.701} &                .669 &                 .698 &            {\bf.712}\phantom{***} \\
pl\_1    &            .681 &                .858 &                 \underline{.870} &            {\bf.875}**\phantom{*} &            .707 &                .844 &                 \underline{.848} &            {\bf.855}*** \\
pl\_2    &            .619 &                .803 &                 \underline{.814} &            {\bf.825}**\phantom{*} &            .697 &                .820 &                 \underline{.834} &            {\bf.839}**\phantom{*} \\
\hline
Avg. &            .688 &                .828 &                 \underline{.832} &            {\bf.844}\phantom{***} &            .757 &                .806 &                 \underline{.831} &            {\bf.839}\phantom{***} \\
\hline
\end{tabular}
\caption{\label{tab:monolingual_average} 
	Results of the monolingual experiment for the WEI baseline, two reference methods (LR and KNN) as well as our FFNN model in Pearson $r$. Best result per data set and emotion format in bold, second best result underlined; significant difference (paired two-tailed $t$-test) over the second best system  marked with ``*'', ``**'', or ``***'' for  $p<.05$, $.01$, or $.001$, respectively.}
\end{subtable}

\bigskip
\begin{subtable}{10cm}
\centering
\small
\begin{tabular}{|l|rrr|rrrrr|}
\hline
{} & Val & Aro & Dom & Joy & Ang & Sad & Fea & Dsg\\
\hline\hline
en\_1 & .969 & .741 & .848 & .962 & .876 & .871 & .873 & .805\\
en\_2 & \cellcolor{lightblue} .964 & \cellcolor{lightblue} .704 & \cellcolor{lightblue} .861 & .942 & .868 & .821 & .860 & .799\\
es\_1 & .974 & .771 & .863 & \cellcolor{lightblue} .957 & \cellcolor{lightred} .854 & \cellcolor{lightred} .833 & \cellcolor{lightred} .869 & \cellcolor{lightred} .752\\
es\_2 & \cellcolor{lightblue} .986 & \cellcolor{lightblue} .828 & \cellcolor{lightred} .720 & \cellcolor{lightblue} .977 & \cellcolor{lightred} .913 & \cellcolor{lightred} .867 & \cellcolor{lightred} .878 & \cellcolor{lightred} .807\\
es\_3 & \cellcolor{lightblue} .915 & \cellcolor{lightred} .692 & --- & \cellcolor{lightblue} .846 & \cellcolor{lightblue} .839 & \cellcolor{lightblue} .857 & \cellcolor{lightblue} .842 & \cellcolor{lightblue} .744\\
de\_1 & .929 & .745 & --- & .894 & .778 & .644 & .785 & .461\\
pl\_1 & \cellcolor{lightblue} .963 & \cellcolor{lightblue} .787 & --- & \cellcolor{lightblue} .946 & \cellcolor{lightblue} .872 & \cellcolor{lightblue} .826 & \cellcolor{lightred} .805 & \cellcolor{lightblue} .826\\
pl\_2 & \cellcolor{lightblue} .947 & \cellcolor{lightblue} .768 & \cellcolor{lightblue} .760 & \cellcolor{lightblue} .935 & \cellcolor{lightblue} .844 & \cellcolor{lightred} .805 & \cellcolor{lightred} .790 & \cellcolor{lightblue} .819\\
\hline
Avg. & .956 & .754 & .810 & .932 & .855 & .816 & .838 & .752\\
\hline
\end{tabular}
\caption{\label{tab:monolingual_individual} 
	Results of the monolingual experiment per affective dimension in Pearson $r$. Color indicates outperforming human SHR (\colorbox{lightblue}{blue}), being outperformed (\colorbox{lightred}{red}) or SHR not being reported (white; ``---'' meaning that the respective variable is not included).
	}
\end{subtable}
\caption{Results of the monolingual experiment.}
\end{wraptable}

%% file: tabs/multilingual.tex
\begin{wraptable}{h}{8.5cm}
\small
\centering
\begin{tabular}{|l|rr|rrrrr|}
\hline
{} & Val & Aro & Joy & Ang & Sad & Fea & Dsg\\
\hline\hline
en\_1 & .966 & .683 & .955 & .858 & .838 & .817 & .781\\
en\_2 & \cellcolor{lightblue} .956 & \cellcolor{lightred} .642 & .934 & .855 & .810 & .791 & .800\\
es\_1 & .973 & .692 & \cellcolor{lightblue} .951 & \cellcolor{lightred} .786 & \cellcolor{lightred} .802 & \cellcolor{lightred} .782 & \cellcolor{lightred} .682\\
es\_2 & \cellcolor{lightblue} .985 & \cellcolor{lightblue} .735 & \cellcolor{lightblue} .974 & \cellcolor{lightred} .881 & \cellcolor{lightred} .860 & \cellcolor{lightred} .835 & \cellcolor{lightred} .787\\
es\_3 & \cellcolor{lightblue} .908 & \cellcolor{lightred} .548 & \cellcolor{lightblue} .839 & \cellcolor{lightblue} .821 & \cellcolor{lightblue} .850 & \cellcolor{lightblue} .807 & \cellcolor{lightred} .728\\
de\_1 & .927 & .708 & .889 & .767 & .618 & .760 & .458\\
pl\_1 & \cellcolor{lightblue} .957 & \cellcolor{lightblue} .666 & \cellcolor{lightblue} .937 & \cellcolor{lightblue} .848 & \cellcolor{lightred} .784 & \cellcolor{lightred} .745 & \cellcolor{lightred} .801\\
pl\_2 & \cellcolor{lightblue} .938 & \cellcolor{lightblue} .720 & \cellcolor{lightblue} .932 & \cellcolor{lightblue} .816 & \cellcolor{lightred} .785 & \cellcolor{lightred} .751 & \cellcolor{lightblue} .809\\
\hline
Avg. & .951 & .674 & .926 & .829 & .793 & .786 & .731\\
\hline
\end{tabular}

\caption{\label{tab:crosslingual}
	Results of crosslingual experiment in Pearson $r$. 
	Color indicates outperforming human SHR (\colorbox{lightblue}{blue}), being outperformed (\colorbox{lightred}{red}) or SHR not being reported (white).
	}
\end{wraptable}

%% file: tabs/constructed.tex
\begin{wraptable}{T}{10cm}
	\small
	\centering
	\begin{tabular}{|llllr|}
		\hline
		Mth & Lng & Format & Source & \#Words \\
		\hline\hline
		m & en & BE5  & \newcite{Warriner13}& 12,884\\
		m & es & VAD & \newcite{Stadthagen17}	&	10,489\\
		m & de & BE5 &  \newcite{Vo09} & 944 \\
		m & pl & BE5 & \newcite{Imbir16} & 3,633 \\
		\hline
		c & it & BE5 & \newcite{Montefinese14} & 1,121 \\
		c & pt & BE5 & \newcite{Soares12} & 1,034 \\
		c & nl & BE5 & \newcite{Moors13} & 4,299 \\
		c & id & BE5 & \newcite{Sianipar16} & 1,487 \\
		c & zh & BE5 & \newcite{Yu16naacl}; \newcite{Yao17} & 3,797 \\
		c & fr & BE5 & \newcite{Monnier14} & 1,031 \\
		c & gr & BE5 & \newcite{Palogiannidi16} & 1,034 \\
		c & fn & BE5 & \newcite{Eilola10} & 210 \\
		c & sv & BE5 & \newcite{Davidson14} & 99 \\
		\hline
	\end{tabular}
	\caption{\label{tab:constructed}
		Overview of automatically constructed emotion lexicons; mapping methodology (\underline{m}onolingual or \underline{c}rosslingual), language (codes according to ISO 639-1), target emotion format, source lexicon of the mapping process and number of previously unknown ratings (excluding those present in other lexicons).}
\end{wraptable}

%% file: conclusion.tex
\section{Conclusion}
\label{sec:conclusion}

In this paper, we addressed the relatively new task of \textit{emotion representation mapping}. It aims at transforming emotion ratings for lexical units from one emotion representation format into another one, e.g., mapping from Valence-Arousal-Dominance representations to Basic Emotion ones. Based on a large-scale evaluation we gathered solid empirical evidence that the proposed neural network model consistently outperforms the previous state-of-the-art performance figures in both word emotion induction and emotion representation mapping. Hence, the approach we propose currently constitutes the best-performing method for automatic emotion lexicon creation.

We also proposed a novel methodology for comparison against human rating capabilities based on normalized split-half reliability scores.
For the first time, this allows for a large-scale evaluation against human performance. Our experimental data suggest that our models perform competitive relative to human assessments, even in cross-lingual applications, thus producing (near) gold quality data. We take this as a strong hint towards the reliability of the methods we propose.

Finally, we used these models to produce new emotion lexicons for 13 typologically diverse languages which are publicly available along with our code and experimental data (see Footnote \ref{fn:github}).